\documentclass[10pt,twocolumn,letterpaper]{article}

\usepackage{cvpr}              

\usepackage{graphicx}
\usepackage{amsmath}
\usepackage{amssymb}
\usepackage{booktabs}

\usepackage{svg}

\usepackage[pagebackref,breaklinks,colorlinks]{hyperref}

\usepackage[capitalize]{cleveref}
\crefname{section}{Sec.}{Secs.}
\Crefname{section}{Section}{Sections}
\Crefname{table}{Table}{Tables}
\crefname{table}{Tab.}{Tabs.}

\def\cvprPaperID{*****} 
\def\confName{Advances in Computer Vision}
\def\confYear{2022}

\begin{document}

\title{UNVEILING SPACES
\\ Architecturally meaningful semantic descriptions from images of interior spaces}

\author{Demircan Tas\\
Massachusetts Institute of Technology\\
77 Massachusetts Ave, Cambridge, MA 02139\\
{\tt\small tasd@mit.edu}
\and
Rohit Priyadarsh Sanatani\\
Massachusetts Institute of Technology\\
77 Massachusetts Ave, Cambridge, MA 02139\\
{\tt\small sanatani@mit.edu}
}
\maketitle

\begin{abstract}
   There has been a growing adoption of computer vision tools and technologies in architectural design workflows over the past decade. Notable use cases include point cloud generation, visual content analysis, and spatial awareness for robotic fabrication. Multiple image classification, object detection, and semantic pixel segmentation models have become popular for the extraction of high-level symbolic descriptions and semantic content from two-dimensional images and videos. However, a major challenge in this regard has been the extraction of high-level architectural structures (walls, floors, ceilings windows etc.) from diverse imagery where parts of these elements are occluded by furniture, people, or other non-architectural elements. This project aims to tackle this problem by proposing models that are capable of extracting architecturally meaningful semantic descriptions from two-dimensional scenes of populated interior spaces. 1000 virtual classrooms are parametrically generated, randomized along key spatial parameters such as length, width, height, and door/window positions. The positions of cameras, and non-architectural visual obstructions (furniture/objects) are also randomized. A Generative Adversarial Network (GAN) for image-to-image translation (Pix2Pix) is trained on synthetically generated rendered images of these enclosures, along with corresponding image abstractions representing high-level architectural structure. The model is then tested on unseen synthetic imagery of new enclosures, and outputs are compared to ground truth using pixel-wise comparison for evaluation. A similar model evaluation is also carried out on photographs of existing indoor enclosures, to measure its performance in real-world settings.
\end{abstract}


\begin{figure}[h]
  \centering
  \includegraphics[width=0.98\linewidth]{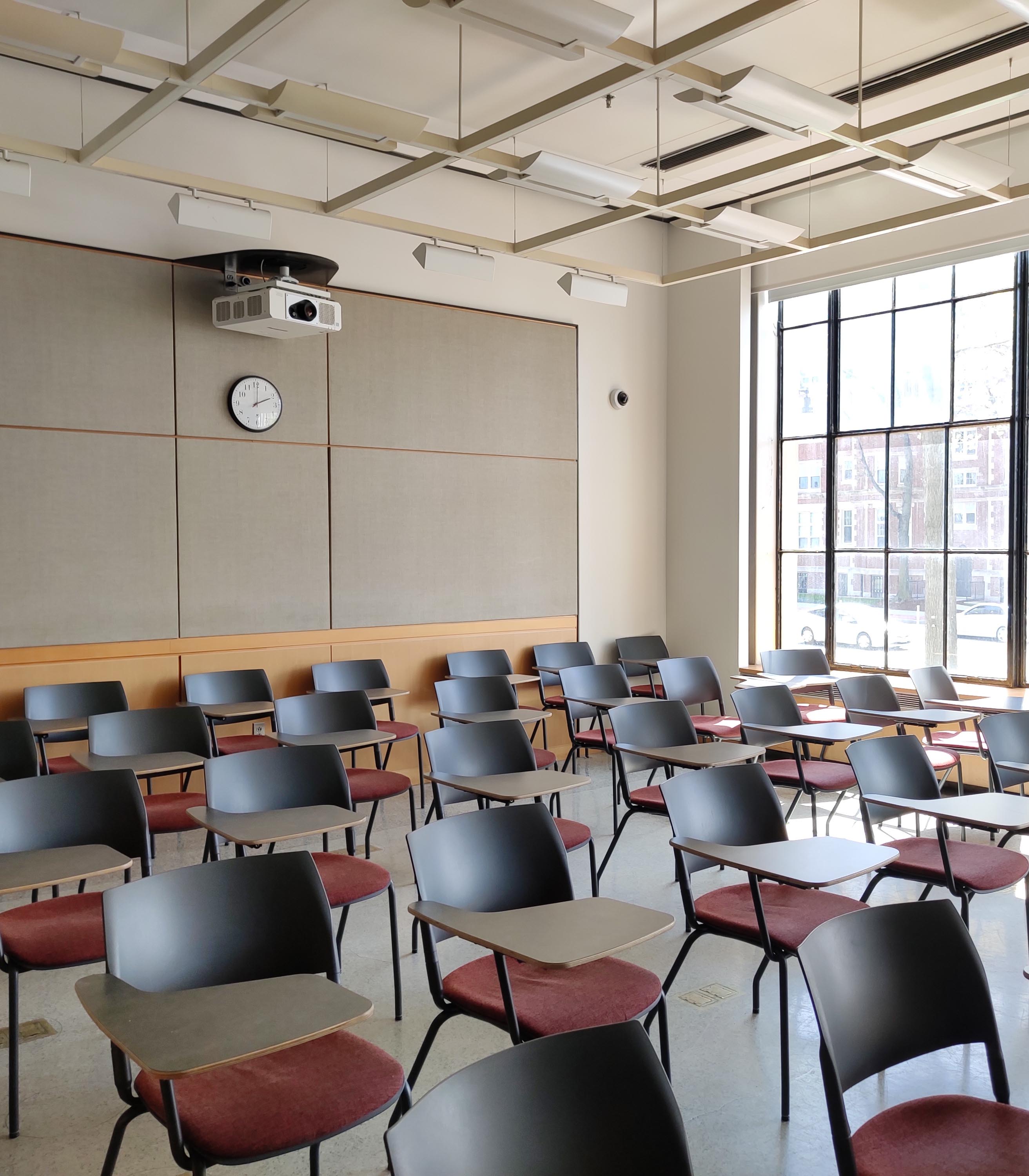}
  \caption{A classroom from MIT, with a variety of objects (clock, projector, wall panels, chairs, etc.) that are irrelevant for an architecture model}
  \label{fig:classroom}
\end{figure}

\section{Collaboration and Logistics}

Presented work is the result of a collaboration between Demircan Tas (DT) and Rohit Priyadarsh Sanatani (RS). RS is mostly responsible for implementing the the Pix2Pix model and domain adaptation, while DT implemented synthetic data generation and data augmentation. However, authors' efforts overlap substantially in terms of developing a holistic model for defining architecturally significant features and eliminating others. System design and research, data collection, documentation, as well as troubleshooting were conducted with equal effort from both authors\footnote{Sections with individual efforts from specific authors are specified with footnotes. Unless explicitly stated, presented work, findings, and comments result from joint efforts.}.

\begin{figure*}[h!t]
  \centering
  \includegraphics[width=\linewidth]{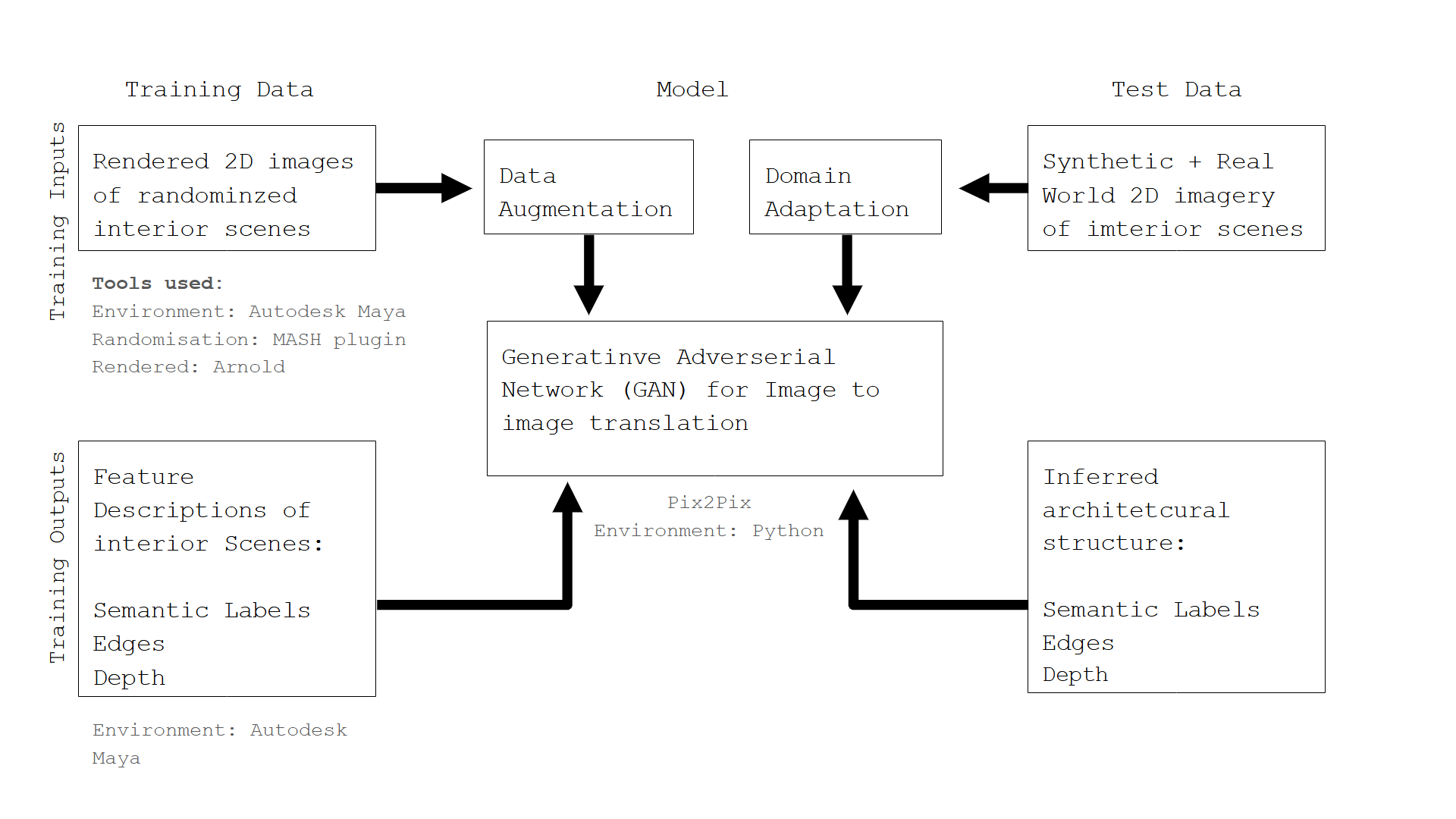}
  \caption{System diagram}
  \label{fig:architecture}
\end{figure*}

\section{Introduction and Background}
\label{sec:intro}

The Architecture, Engineering and Construction (AEC) industry has been increasingly relying on recent advances in computer vision for a variety of workflows. While photogrammetric three-dimensional point cloud generation from multiple shots of interior or exterior scenes has been around for quite a few years within the domain of heritage/as-built documentation, generating point clouds from single scenes using monocular depth prediction is becoming increasingly popular \cite{yuniarti2019review}, \cite{madhuanand2021self}. Object detection models play important roles in the analysis of behavioral patterns within the built environment \cite{sreenu2019intelligent}. There have been notable developments within the field of robotic construction, fabrication and assembly, aiding efforts to provide robotic agents with spatial awareness \cite{buchli2018digital}.

A central theme in computer vision workflows is the extraction of high-level semantic content and symbolic descriptions from low-level descriptions (pixels) in images and videos. Multiple image classification, object detection, and semantic segmentation models have been used in the recent past for this purpose. Datasets such as ImageNet \cite{deng2009imagenet} and Places \cite{zhou2017places} allow for training models to classify scenes into symbolically meaningful categories. Moreover, datasets such as ADE20K \cite{zhou2017scene} and CityScapes \cite{cordts2016cityscapes} allow for the segmentation of each pixel of an indoor or outdoor image, into semantic categories.

While these models are suitable for a variety of use cases, they are not adequately equipped to extract high level architectural structure from indoor images. The Building Information Modeling (BIM) paradigm has over the years developed a hierarchical framework of architectural descriptions. These descriptions consist of basic concepts such as walls, floors, ceilings and roofs, along with nested classes of objects such as doors, windows and skylights that are hosted within these units. While conventional object detection/segmentation models are able to classify each pixel as a ceiling or as a wall, they are not able to extract the high-level structure of these units, or the relationships among them. This is largely due to the fact the descriptions encoded within the training datasets of these models do not contain adequate information with regards to the hierarchical structuring of architectural elements. As a result, these models lack the ability to infer the structure of architectural elements which are occluded by visual obstructions that are natural in everyday interior scenes.

\begin{figure*}[]
  \centering
  \includegraphics[width=0.98\linewidth]{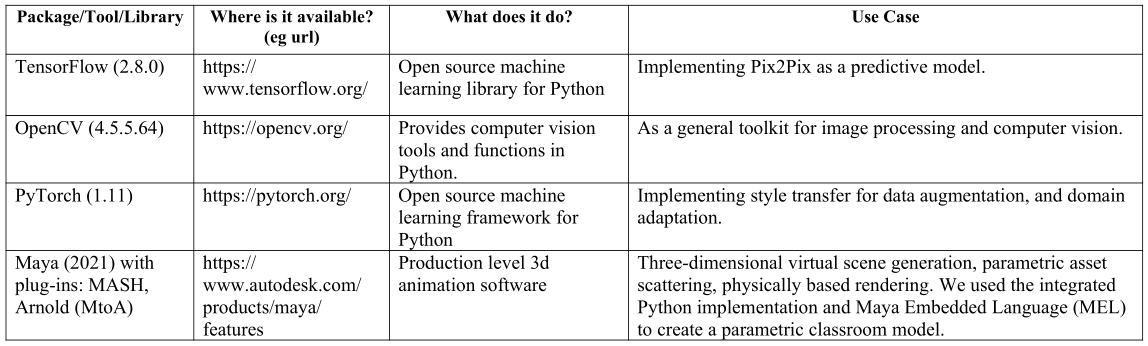}
  \caption{Software and libraries used for implementation.}
  \label{fig:classroom}
\end{figure*}

\begin{figure*}[]
  \centering
  \includegraphics[width=0.98\linewidth]{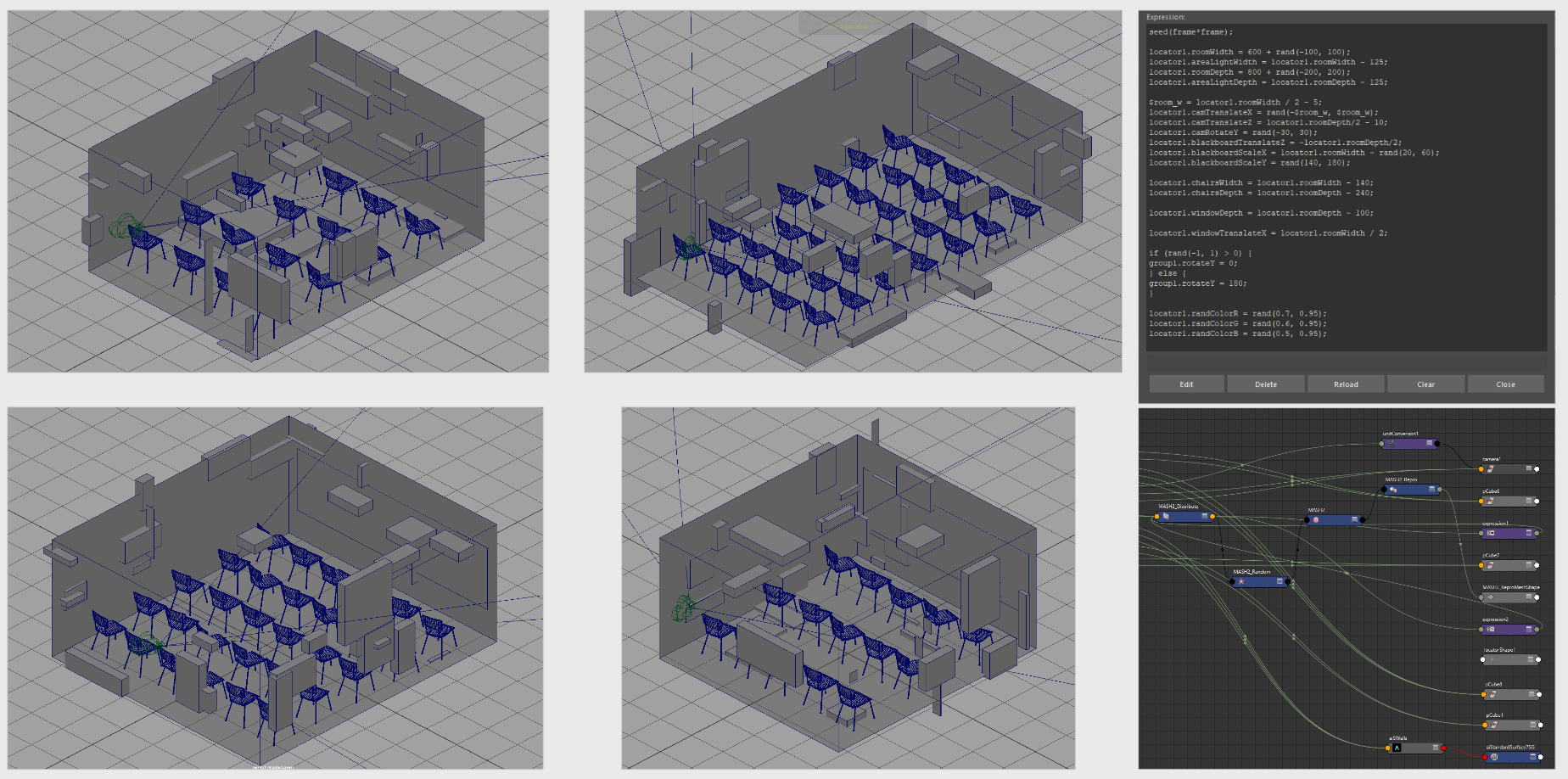}
  \caption{Parametric Maya scene, MEL expressions for randomization, and node editor connections}
  \label{fig:mscene}
\end{figure*}

\subsection{Related Work\: Generative Adversarial Networks}

Recent advances in the development of robust Generative Adversarial Networks (GAN's) \cite{creswell2018generative} show a lot of promise with regards to the task of inferring missing/occluded portions of images. Training of GAN's involve the training of two networks pitted against each other. One of the networks - the generator - is successively trained to generate images that match the distribution of training samples. The other network - the discriminator - is successively trained to discriminate between samples generated by the generator, and those drawn from the training data distribution. Through iterative training involving these two adversarial networks, GAN's allow for the learning of complex representations of real world topologies even in the absence of large amount of labeled training data. 

GAN's have gained popularity in the recent past for a variety of generative tasks. Notable examples include the generation of non-existent human faces, pets, or even cities. Models such as StyleGAN \cite{karras2019style} have become important in these regard. The use of GAN's for image to image translations can be useful for inferring high-level topological structure from low-level representations. Models such as Pix2Pix \cite{isola2017image}, shorthand for 'pixel to pixel' have proven to be powerful and flexible tools for numerous research methodologies.    

This project draws upon the potential of GAN's to propose models that are capable of extracting architecturally meaningful semantic descriptions from two-dimensional scenes of populated interior spaces. It uses image-to-image translation through a GAN trained on a dataset comprising of architecturally meaningful semantic abstractions, in order to extract high level architectural structure from low level pixel representations of interior scenes.

\section{Methodology}

\subsection{Generating synthetic data for training}

\footnote{This section presents efforts of Demircan Tas}1000 parametrically defined interior scene geometries were generated in Maya, based on photographs of classrooms from MIT. Classrooms were chosen as a context due to the relatively low variety of furniture and ease of access. While we used a generic model for chairs, all remaining items were represented by cube geometries. A black surface was explicitly defined for each room as a blackboard. The remaining cubes were scattered randomly.

Box geometries with randomized \textit{width}, \textit{height}, and \textit{depth} values were scattered on walls, floors and ceilings via randomized $u, v$ coordinates using \textit{MASH} plug-in for Maya. Moreover, diffuse colors of box geometries were randomized in \textit{hue}, \textit{saturation}, and \textit{value}.

Each of these scenes are rendered from a camera with random rotations in the $y$ axis and random translations in the $x$ (lateral). Camera yaw was constrained within $\pm 30$ degrees, while pitch was strictly constrained to the horizontal axis, and no camera roll was allowed. Lateral movement was limited between the walls of the room with a 50cm offset. Each camera angle was used to generate a 512x512 pixel photo realistic image, with physically-based material and lighting information.

For each shot as generated above, output features containing semantic information of architectural structure were generated. These were encoded as 3 output images per scene, namely: Walls, ceilings and floors (defined as red, green, and blue), and depth maps. A total of 1000 images and corresponding feature sets were generated for training.

Given our render resolution, boxes were visually convincing placeholders for a variety of objects from the real classrooms. However, this approach is limited to low resolutions and objects that are not close to the camera. For a higher fidelity model, a library of objects will be necessary with matching physically based materials.

\begin{figure}[]
  \centering
  \includegraphics[width=0.49 \linewidth]{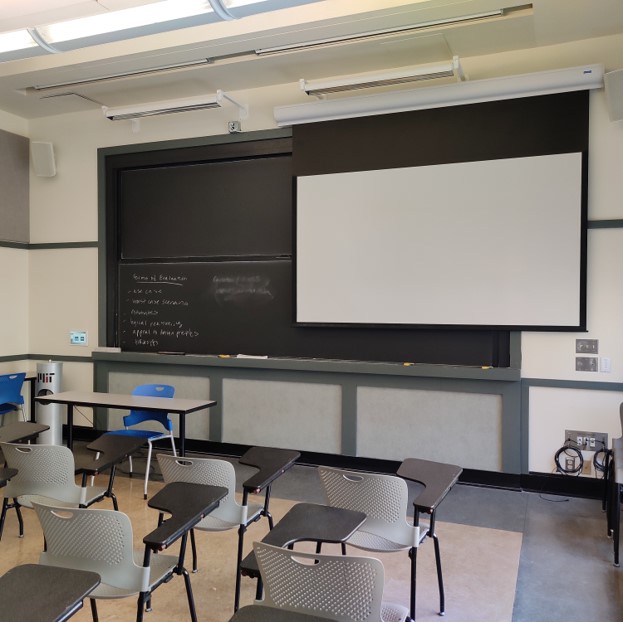}
  \includegraphics[width=0.49 \linewidth]{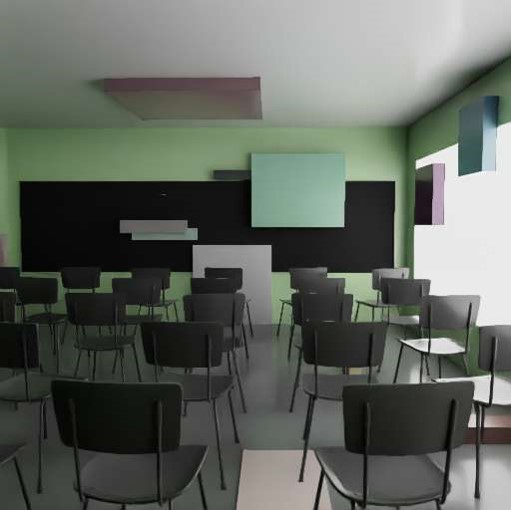} \\
  \includegraphics[width=1.0 \linewidth]{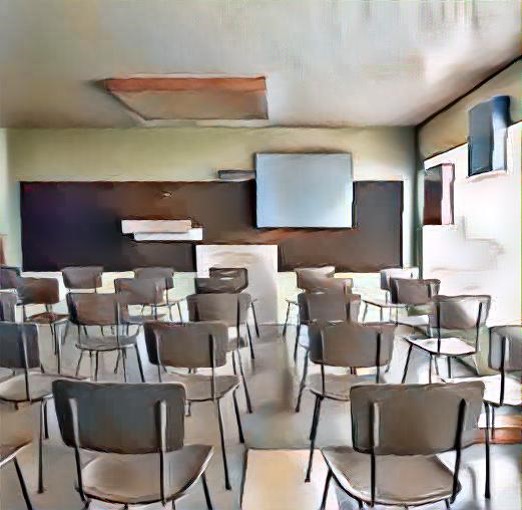}

  \caption{Classroom photo (top left), rendered image (top right), augmented synthetic image (bottom)}
  \label{fig:dataAug}
\end{figure}

\subsubsection{Data Augmentation}

\footnote{This section presents efforts of Demircan Tas}To improve the robustness of our model, we applied a style transfer algorithm based on \cite{gatys2016image}, stating that style and content representations can be separated in a convolutional neural network (CNN). Given a target and input image, style transfer migrates low level features of the target image into the input image while preserving high level features or \textit{semantic} qualities of the input image. The output is an image where objects' shapes (or semantics) are preserved, but color, texture, and border qualities are taken from classroom photos \ref{fig:dataAug}.

Using style transfer, we augment synthetic datasets to better represent the target context. Moreover, we use it as a means of domain randomization to account for the differences between synthetic objects (random cubes) and real ones (panels, projectors, key switches, screens, etc.).

\subsection{Supervised training on synthetic data}

\footnote{This section presents efforts of Rohit Priyadarsh Sanatani}A Pix2Pix image-to-image translation model \cite{isola2017image} was trained on the image pairs. Pix2Pix has been used in the past for a variety of image-to-image translation tasks, and has proven successful and robust in generating visual content from high level topological representations. Notable examples include the generation of photo-realistic images from conceptual sketches, or the generation of street level urban imagery from semantically segmented images. More importantly, the model has also been successful in generating/extracting high level representations from noisy two dimensional imagery. Examples in this regard include the extraction of building footprint and street network information from satellite imagery of urban environments.

As discussed earlier, our proposal involves the extraction of high level, architecturally meaningful semantic representations from two-dimensional imagery of interior scenes. We note that both the synthetic interior scenes, as well as their architectural representations (labels and depth) are raster pixel-based images. However, since the output pixels embody architecturally meaningful information (the class of architectural elements and spatial relations between them) and missing parts are filled by Pix2Pix, the challenge posed by occlusion in the reconstruction of three-dimensional architectural structure from two-dimensional populated imagery is overcome.

For training, input-output pairs of images were pre-processed and down-scaled to 256x512 pairs. They were then split up into training (750), validation (200) and testing (50) sets. In addition to the test samples in the dataset, 30 real images of MIT classrooms were collected and used as an additional test set to evaluate the performance of the model in real world scenarios. Training parameters for Pix2Pix were taken into consideration, and were finalised as \texttt{batch\_size = 1} and \texttt{lambda = 100}. 

In total, three models were trained as part of our methodology. The first model \textbf{($M_{1}$)} was trained to predict class labels from synthetic imagery without any data augmentation. The second model \textbf{($M_{2}$)} was trained on synthetic imagery which was augmented using style transfer to match the distribution of the real world imagery as discussed earlier. Finally, the third model \textbf{($M_{3}$)} was trained to predict depth from synthetic imagery without style transfer. 

Training for \textit{($M_{1}$)} was first carried out over 1000 steps to gauge the performance of the model with current settings, and then continued successively over 10,000, 20,000 and finally 30,000 steps. Predictions on the test set were saved for each of the three training benchmarks. Training time was around 1 hour. For \textit{($M_{2}$)}, training was carried out over 10,000 steps. \textit{($M_{3}$)} was trained on the depth labels also over 10,000 steps.

\begin{figure}[b]
  \centering
  \includegraphics[width=1.0 \linewidth]{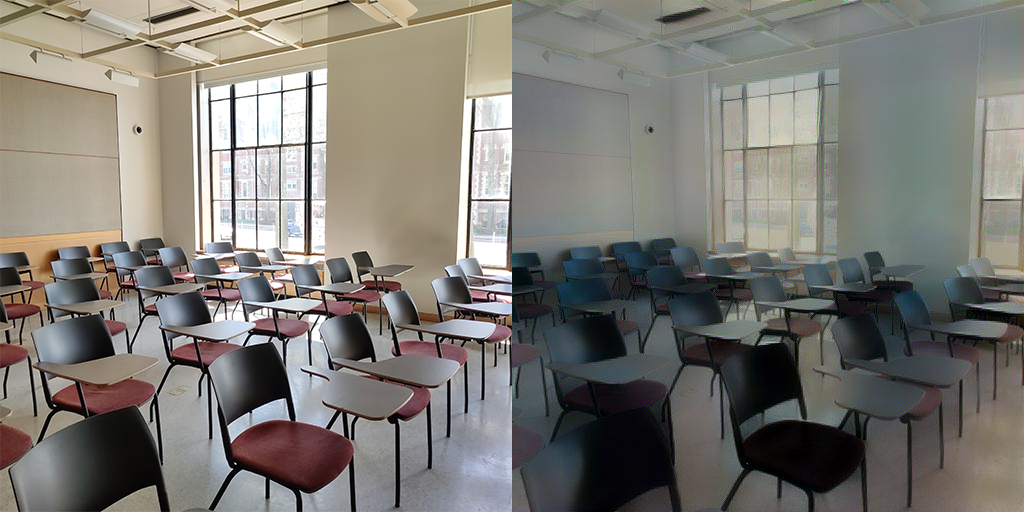}
  \caption{Real photograph (left) adapted styled to match synthetic imagery (right)}
  \label{fig: domain_adapt}
\end{figure}

\begin{figure*}[]
  \centering
    \includegraphics[width=0.95\linewidth]{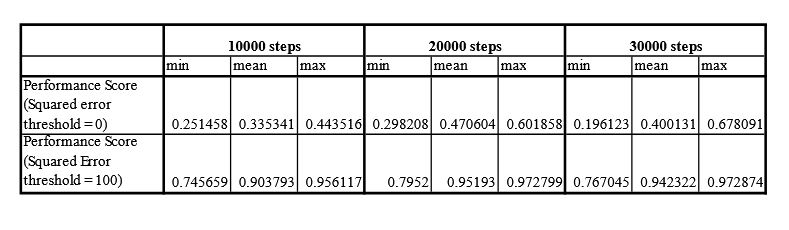}
  \caption{Model performance}
  \label{fig: M1_acc_table}
\end{figure*}

\subsubsection{Domain Adaptation}

\footnote{This section presents efforts of Rohit Priyadarsh Sanatani}While data augmentation provides visually consistent images that yield marginally better results with the test data, predictions on photographs was not satisfactory. As an alternative approach to data augmentation, we also used style transfer using synthetic images for style, and test photos for content, prior to running the prediction model. This technique was used to match the the domain (real world pictures) to the data (synthetic images).

Using this approach, we sacrifice a minimal amount of accuracy on test data from synthetic images, but achieve superior accuracy in predictions on real world images.

\section{Model Performance and Evaluation}

\begin{figure}[]
  \centering
  \includegraphics[width=1.0 \linewidth]{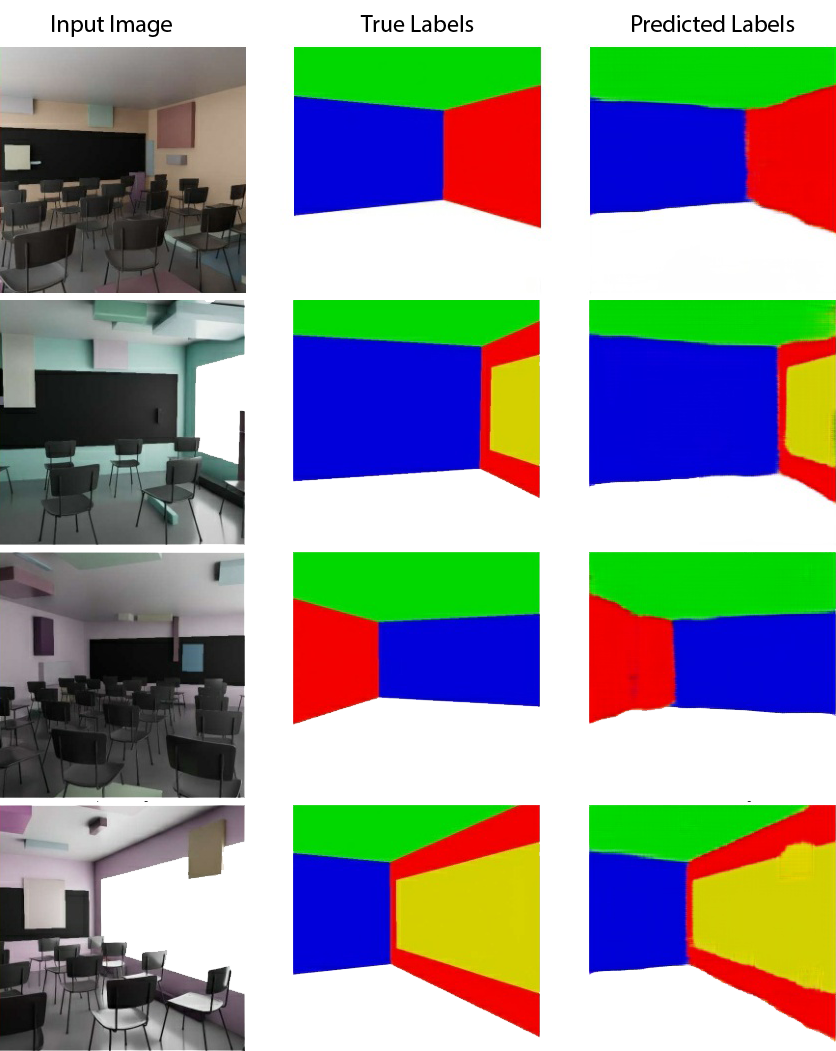}
  \caption{Sample predictions for Model MI on synthetic test data}
  \label{fig: MIperf}
\end{figure}

\subsection{Evaluation Metrics}

For evaluation of the model, a pixel-wise comparison of predicted imagery and ground truth was undertaken. Two different methods were tested out in this regard. The first method involved involved iterating over each pixel of the predicted image, and checking equality with the corresponding pixel in the ground truth. The evaluation metric is defined as the percentage of pixels in the predicted image that match the ground truth. This method yields extremely low evaluation scores. This is because, while the two images may appear to be very similar upon visual comparison, and even have pixel values that are very similar to each other, the values have to be exactly the same to register as correct\footnote{Pix2Pix yields subtle differences in predicted colors. A \textit{red} in the prediction, while perfectly red to the observer, is usually off by a slight margin in terms of pixel values.}.

To work around this, the second evaluation metric considers a threshold value to check for equality between the pixels of the two images. If the squared error between the values lay within the threshold, they are considered equal. Choosing an appropriate threshold is critical in this regard. A threshold that is too large would fail to detect incorrect label predictions. A very low threshold on the other hand would penalize predictions that are very close to the ground truth and lead to low accuracy scores. After iterative testing, a threshold value of $10$ was decided upon ($10\times10 = 100$ for comparison of squared error).

\subsection{Model Performance}

\subsubsection{Model $M_{1}$\: Trained on Synthetic Data without Augmentation}

Model $M_{1}$ yielded impressive results in predicting architectural labels from synthetic interior scenes. Specifically, the structure of edges, and window boundaries were accurately reconstructed. Performance scores varied between 0.792 and 0.956 over the 50 test samples, with a score of 0.904 (when trained through 20,000 steps). When trained for 30,000 steps however, the prediction accuracy was lower, indicating that the model over-fitted. Table \ref{fig: M1_acc_table} below depicts the performance scores at different stages of training. Fig. \ref{fig: MIperf} shows representative predictions, as well as notable failure cases.

\subsubsection{Model $M_{2}$\: Trained on Synthetic Data with Data Augmentation}

\begin{figure}[]
  \centering
  \includegraphics[width=1.0 \linewidth]{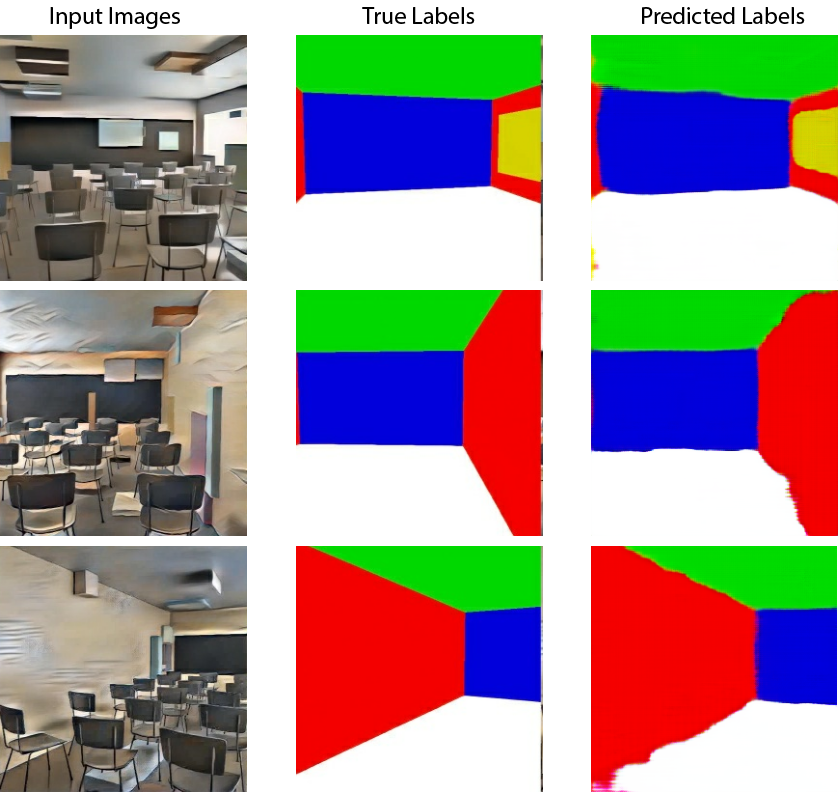}
  \caption{Sample predictions for Model $M_{2}$ on synthetic and augmented test data}
  \label{fig: M2perf}
\end{figure}

For model $M_{2}$, training was carried out on augmented data using style transfer as discussed earlier. This was done to improve the robustness of the model, and evaluate its performance not only on synthetic data, but also on real world imagery of actual classrooms. This was also done to provide a comparative evaluation of the system's performance when trained on augmented data and tested on real world imagery, as opposed to being trained on synthetic data and tested in adapted test imagery. Figure \ref{fig: M2perf} depicts examples of the model's performance when tested on augmented synthetic data. The mean performance score (with squared error threshold = 100) was 0.940, with minimum and maximum scores of 0.879 and 0.959 respectively. 

\begin{figure}[]
  \centering
  \includegraphics[width=1.0 \linewidth]{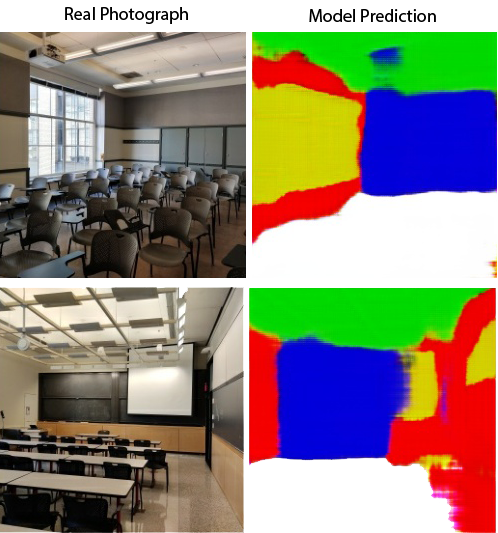}
  \caption{Sample predictions for Model $M_{2}$ on real world imagery}
  \label{fig: M2real}
\end{figure}

When tested on real world imagery, however, the performance of the model was far worse than expected. There were multiple failure cases, with many of the failures exhibiting systematic and consistent patterns. Figure \ref{fig: M2real} shows some representative predictions. When the same real world images were adapted to match the distribution of synthetic data and predicted through model $M_{1}$, the results turned out to be far better. These are discussed further in the proceeding section.

\subsubsection{Model $M_{3}$: Trained on Synthetic Data and Depth Maps}

\begin{figure}[]
  \centering
  \includegraphics[width=1.0 \linewidth]{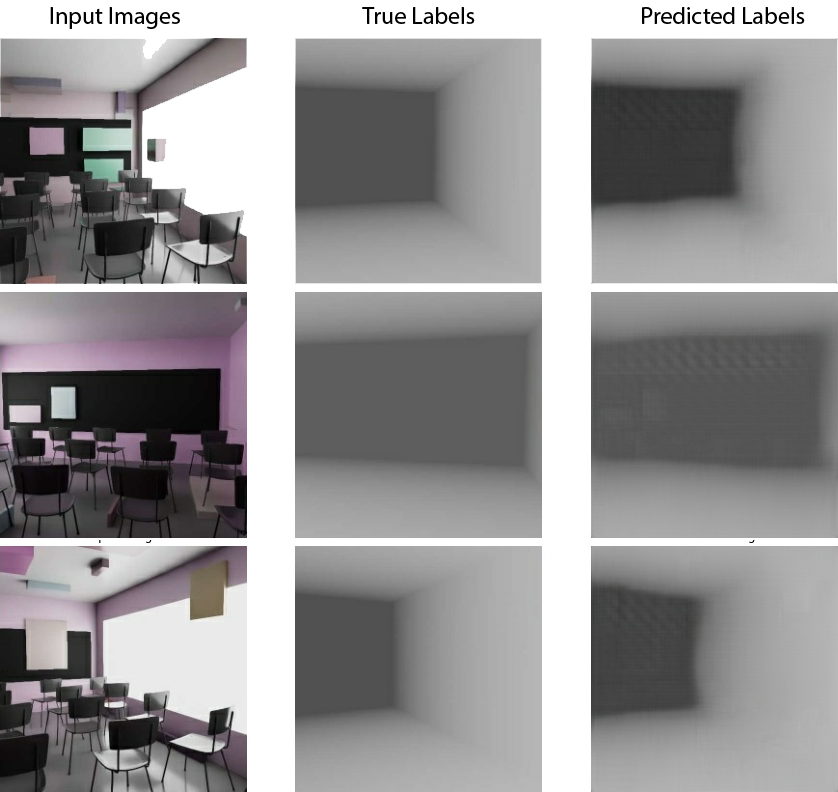}
  \caption{Sample predictions for Model $M_{3}$ on synthetic imagery}
  \label{fig: M3pred}
\end{figure}

Model $M_{3}$ was trained on synthetic data without augmentation, and the corresponding depth-maps of the enclosures without any non-architectural elements. Predicting depth of architectural elements at each point pixel of an enclosure irrespective of occlusions is critical, as it allows for the three-dimensional reconstruction of architectural features, when coupled with label predictions. The model performance seemed impressive visually, but the evaluation metrics were worse than that of models $M_{1}$ and $M_{2}$. The performance scores ranged between 0.30 and 0.97 with an average performance score of 0.79. Depth images are single channel, and unlike the label images, each pixel denotes a continuous depth variable rather than a discrete class. As a result, minor variations in the value of the pixels correspond to significant errors in depth prediction. Figure \ref{fig: M3pred} shows representative examples of depth prediction.

\section{Discussion}

\subsection{Notable failure cases and patterns}

\begin{figure}[t]
  \centering
  \includegraphics[width=1.0 \linewidth]{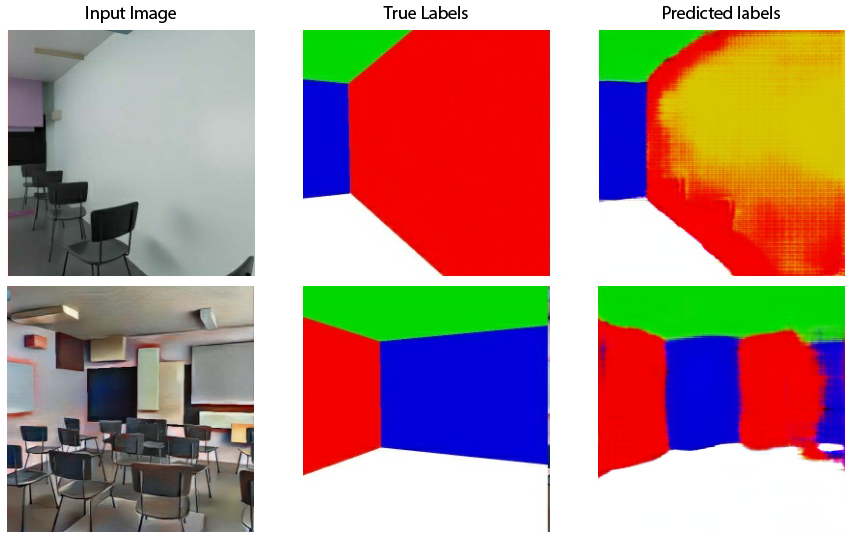}
  \caption{Notable failure patterns}
  \label{fig: failures}
\end{figure}

While both models performed well on the synthetic test data which matched the training distribution, there were nevertheless notable cases of failure. In general, two major patterns of failure consistently occured across most failure samples. The first pattern involved failure to accurately detect window boundaries, or detecting windows where there were none. This was because, the windows in the synthetic images were characterized by very high pixel intensities. This resulted in parts of walls which also exhibited high intensities to be mis-labeled as windows (Fig \ref{fig: failures} (top)). We see this pattern repeating in many cases, with very bright patches of walls being classified as windows.

The second failure pattern involved the detection of wall edges. In many cases, especially with augmented training data, edges of obstructions were misinterpreted by the model to be corners between two walls. This resulted in single walls with obstructions being labeled as two or more walls separated by the obstruction edges. 

Fig. \ref{fig: M2real} also betrays notable failure patterns in predictions from real world imagery. In the upper image, window prediction (yellow) \textit{bleeds} into the side walls (red) due to the continuity of horizontal line features, with transoms\footnote{Horizontal crossbars in windows, over doors, or between windows, doors, or fanlights\cite{webster2006merriam}}. An artifact of this effect can be observed in the wavy vertical transition between window and side wall predictions. The bottom image presents an error case where high intensity areas of the right wall are mis-classified as windows. This issue is caused by window features in our test images consisting of over-exposed areas.

\subsection{Data augmentation vs Domain adaptation}

\begin{figure}[t]
  \centering
  \includegraphics[width=1.0 \linewidth]{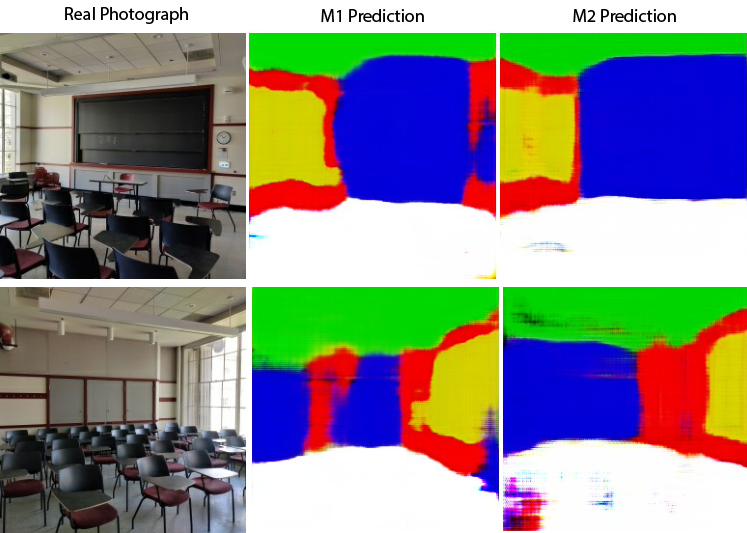}
  \caption{Performance comparison of data augmentation ($M_{1}$) and domain adaptation ($M_{2}$) on real world imagery }
  \label{fig: comparison}
\end{figure}

As we see from the results, model $M_{1}$ trained on augmented data yields better results on the synthetic augmented test data (mean performance score = 0.95), but fails to perform well when tested on real world imagery. On the other hand, model $M_{2}$ performs equally well on synthetic un-augmented test images (mean performance score = 0.94), but drastically better results with real-world images with domain transfer. Fig \ref{fig: comparison} illustrates this point, and compares the performance of both models on real world imagery. We thus conclude that training a model on un-augmented images and testing them on real world imagery adapted to match the training distribution shows promise, and can be taken forward for further development of this methodology. 

\subsection{Limitations}

The results, though impressive, highlight a couple of important limitations of this methodology. Firstly, despite the use of data augmentation and domain adaptation to improve the model performances to real world data, a qualitative assessment of the outcomes reveal that they perform far worse on real world imagery as compared to the synthetic imagery that they were trained on. This however also opens up potentials for improving upon this work through the training of image-to-image GAN's on labeled real world data, rather than synthetic data. Secondly, the scope of this work was restricted to a certain architectural typology, namely classrooms. Applying this model to other typologies will result in lower prediction accuracies. A randomised generative algorithm similar to the one used in this work can be created for other typologies as well, to train the model to be more robust against a wider distribution of spatial typologies. Finally, while our work aimed at extracting both semantic labels as well as depth, the methodology can be extended to generate three-dimensional point clouds and meshes directly from two-dimensional imagery.

\section{Conclusion}

The body of research described in this paper is one of the first lines of inquiry that aim to extract architectural structure from noisy two-dimensional imagery where non-architectural obstructions occlude significant parts of the underlying architectural elements. Our results show that an image-to-image translation methodology is effective in this regard, capable of extracting high level structure from both synthetic as well as real world imagery.

As discussed earlier, a major challenge in contemporary architectural photogrammetry tasks is the semantic enrichment of photogrammetric meshes to reflect true architectural labels. This is valuable form a Building Information Modeling (BIM) point of view, as models generated through such a method will reflect the vocabulary of BIM rather than that of geometric modeling. Moreover, current photogrammetric models are singular meshes, where architectural elements such as walls and floor remain fused with non architectural elements such as furniture. A methodology for generating purely architectural models using photogrammetry is valuable for the AEC industry, with the potential to greatly speed up the generation of BIM models of existing structures. 

We anticipate that the methodology described in this paper will prove to be a valuable foray into a variety of use cases within the world of architectural photogrammetry, building information modeling, and heritage documentation.




{\small
\bibliographystyle{ieee_fullname}
\bibliography{egbib}
}

\end{document}